# The Effects of Age, Gender and Region on Non-standard Linguistic Variation in Online Social Networks

Claudia Peersman, Walter Daelemans, Reinhild Vandekerckhove,
Bram Vandekerckhove, and Leona Van Vaerenbergh.

CLiPS, Department of Linguistics, University of Antwerp, Belgium

**Abstract**

We present a corpus-based analysis of the effects of age, gender and region of origin on the production of both 'netspeak' or 'chatspeak' features and regional speech features in Flemish Dutch posts that were collected from a Belgian online social network platform. The present study shows that combining quantitative and qualitative approaches is essential for understanding non-standard linguistic variation in a CMC corpus. It also presents a methodology that enables the systematic study of this variation by including all non-standard words in the corpus. The analyses resulted in a convincing illustration of the Adolescent Peak Principle. In addition, our approach revealed an intriguing correlation between the use of regional speech features and chatspeak features.

# Introduction

In linguistics, non-standard language usage is defined as any language usage that differs from the (un)officially recognized prestige language variant as it is used primarily in written language and formal speech situations (Trotta, 2011). Correlations between the use of non-standard language and sociological variables such as age and gender have already been observed in many spoken discourse studies. In her overview of variationist sociolinguistic research, Tagliamonte (2012: 32) states that there is a consensus among sociolinguists (e.g., Wolfram, 1969; Labov, 1972; Trudgill, 1983; Milroy, Milroy, Hartley & Walshaw, 1994; Cheshire, 2002) that of "all the sociolinguistic principles, the clearest and most consistent one is the contrast between women and men", which is defined as the Gender Effect. More specifically, (i) in stable sociolinguistic stratification, men tend to use more non-standard forms than women do and (ii) women are usually the innovators in linguistic change (Labov, 1990). With regard to Age Grading, research has shown that people of different ages use speech appropriate to their age group (e.g., Downes, 1984; Labov, 1994; Wardaugh, 2002). That is, when a linguistic variety is not part of the standard language, its usage tends to peak during adolescence (i.e., 15-17 year old), "when peer pressure not to conform to society's norms is greatest" (Holmes, 1992), while pre- and post-adolescents are found to use these variables less frequently (e.g., Labov, 2001; Chambers, 2003). This effect is usually referred to as the Adolescent Peak Principle. However, as social pressure increases and the use of standard language becomes more important for e.g., building a career and/or raising children, people are more inclined to adapt to society's norms. Hence, the use of standard (or prestige) forms tends to peak between the ages of 30 and 55 (e.g., Tagliamonte, 2012).

Although there is no one-to-one relationship between CMC and spoken discourse (e.g., Daft and Lengal, 1984; Kiesler, Siegel & McGuire, 1984; Herring, Stein & Virtanen, 2013), paralinguistic and nonverbal cues, which are absent from the written repertoire, are often compensated by 'netspeak' or 'chatspeak' features, such as emoticons, character flooding and the use of uppercase to express emphasis (e.g., Crystal, 2001). Moreover, the loose cross-turn relatedness in multiparticipant CMC has not only encouraged language play (Herring et al., 2013), leading to the rise of typical chatspeak abbreviations and concatenations (e.g., 'bff' (*best friends* *forever*), 'brb' (*be right back*)), it has also led to a tradeoff between strictly applying spelling rules and maximizing one's typing speed. For example, errors and typos are seldom corrected and punctuation marks are often left out. Although such features are becoming more common across different cultures and languages, Thurlow and Poff's (2013) research showed less convergence of such practices than predicted by Baron (1984). Moreover, they found a tendency to represent 'regiolectal spellings' in text messages (SMS) within the US (see also Eisenstein, O'Connor, Smith & Xing, 2010; Siebenhaar, 2006).

In this study, we introduce a corpus of Flemish Dutch posts from the Netlog social network[1] that was collected without any researchers' interventions, together with the users' profile information (age, gender and location). What makes this corpus interesting for studies on non-standard linguistic variation, is that it contains two types of non-standard language use to which the above mentioned Age Grading and Gender Effect could apply: (i) newly incoming non-standard forms — of which (female) adolescents have been shown to be the innovators in spoken discourse (e.g., Tagliamonte, 2012) — that are characteristic of CMC, and (ii) written representations of regional and dialect forms that are typical for colloquial speech in Flanders (see also Vandekerckhove & Nobels, 2010). However, studies on dialect vitality in Flanders (e.g., Vandekerckhove, 2009; Van Keymeulen, 1993) have shown that especially younger people tend to use fewer dialect forms than they did a few decades ago. Therefore, we examine the effect of age, gender and region on both these types of non-standard language use based on a subset of the Netlog corpus, which was balanced according to age (13 to 49), gender and dialect region (West-Flanders, East-Flanders, Brabant and Limburg). We first perform a qualitative analysis by manually categorizing all non-standard words that were present in this subset as either typical chatspeak features or belonging to one (or more) of the Flemish dialects or regiolects. Based on this linguistic information we set up a forward stepwise mixed-effects logistic regression analysis (Breslow & Clayton, 1993), in which we examine the effects of age, gender and region on the production of both chatspeak and regiolectal forms in our subset. This corpus-based approach, in which we include all occurrences of non-standard language, also enables us to construct a more systematic view of these effects in our CMC corpus.

We start with a description of the Netlog corpus, in which we also illustrate both types of non-standard language use we encountered in it. Next, we turn to the first part of our study in which we describe the compilation of the Netlog subset and elaborate on the operationalization of non-standard language. We also explain which parameters were used to attribute the words to one of both types of non-standard language use. Next, we present the results of the statistical analyses and compare them to previous findings in related research. We conclude our study with an overview of our results and our research contributions.

## The Netlog Corpus

### Structure

Netlog is a Belgian online social networking platform that focuses mainly on European adolescents and has over 99 million members, utilizing over 40 different languages. Members can create a profile page containing blogs, pictures, videos, events, playlists, etc. that can be shared with other members. The entire Netlog corpus contains 1,537,283 Flemish Dutch Netlog posts with a total of 18,713,627 tokens (i.e., words, emoticons and punctuation marks). For each post we were able to obtain information about the age, gender and geographical location of the authors. The posts have an average length of 12.2 word tokens. Table 1 provides an overview of the distribution of the users per main dialect region, gender and age group. Since it was impossible to perform manual verification of the linked profiles, it is possible that some of the profiles contain false information. However, we assume the number of these posts to be too limited to be of concern for the statistical analyses.

*Table 1. Number of users per Flemish region, age group and gender in the Netlog corpus.*

| Age Group | West-Flanders | | East-Flanders | | Brabant | | Limburg | |
|---|---|---|---|---|---|---|---|---|
| | *Male* | *Female* | *Male* | *Female* | *Male* | *Female* | *Male* | *Female* |
| 10s | 7187 | 12,341 | 9366 | 14,763 | 16,964 | 26,701 | 3677 | 6594 |
| 20s | 1799 | 1238 | 2771 | 1764 | 4207 | 2612 | 1121 | 761 |
| 30s | 740 | 572 | 1021 | 705 | 1392 | 890 | 466 | 339 |
| 40s | 726 | 640 | 924 | 846 | 1226 | 1026 | 520 | 385 |
| +50s | 601 | 588 | 783 | 645 | 1162 | 1108 | 359 | 319 |
| **Subtotal** | 14,865 | 18,723 | 11,053 | 15,379 | 24,951 | 32,337 | 6143 | 8398 |
| **Total** | 33,588 | | 26,432 | | 57,288 | | 14,541 | |

**Characteristics of Non-Standard Language Variation in Flanders and its Reflection in Chatspeak**

All of the social network posts in the present corpus contain varieties of Flemish Dutch. These varieties constitute a vertical and horizontal continuum (cf. Auer & Hinskens, 1996). The so-called horizontal continuum relates to the dialectal and regiolectal geolinguistic variation, whereas the vertical continuum ranges from small-scale local dialects to the standard language, in this case Belgian Standard Dutch, which deviates from Netherlandic Dutch in some minor respects. The diaglossic vertical continuum (see also Auer, 2005) is marked by intermediate varieties that are typical of Flemish colloquial speech. Their use seems to be expanding at the expense of the two poles of the continuum (Van Hoof & Vandekerckhove, 2013; Grondelaers & Van Hout, 2011; Vandekerckhove, 2009). These popular 'in-between' or regiolectal (rather than dialectal) varieties are still marked by regional differences (e.g., Geeraerts, 2001), and can be grouped according to their geographical distribution into four main regiolect areas: Brabant (Antwerp and Flemish-Brabant), West-Flanders, East-Flanders and Limburg (see Figure 1). In Table 2 we provide an example post for each region from our corpus and its equivalents in Standard Dutch and English. Since these regional non-standard varieties are frequently used in Flemish Netlog users' posts, they are crucial for the present study. However, a complicating factor is that there is no standard spelling for regional features. As a consequence, non-standard forms can be represented by multiple spelling variants. For example, the non-standard Flemish use of 'schoon' in the meaning of *beautiful*[2] occurs in the Netlog corpus as 'schoon', 'schuun', 'sgoown', 'sgoon', 'skone', 'skoon', 'skwone', 'skwune', 'skwoane', 'skwunne' and 'skwnee'. This enormous amount of orthographic variation proves a major challenge in the selection and categorization of written chat data.

Apart from these occurrences of regional non-standard variation, a second type of non-standard language variation, typical of chatspeak in general, is found in our corpus: Netlog users often omit letters or even entire words or use abbreviations and acronyms in order to maximize their typing speed. In addition, spelling errors are rarely corrected and punctuation marks are often left out. Moreover, to emphasize the content, a flooding of characters or uppercase is often used and (parts of) sentences are concatenated. This is illustrated in Table 3. In certain cases, chatspeak features and non-standard regional features may be combined in one single lexeme. For example, the word 'wroem' is an abbreviated form of 'woaroem', which represents the dialect pronunciation of 'waarom' (*why*) in Brabant, while 'skwne' is short for 'skwunne' or 'skwoane', which is the south-eastern West-Flemish variant of 'schoon' (*beautiful*).

*Figure 1. The Four Main Dialect Regions in Flanders: West-Flanders, East-Flanders, Brabant and Limburg.*

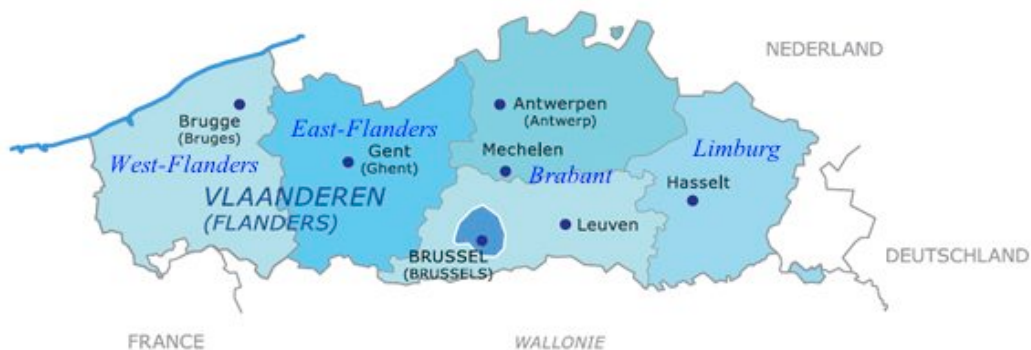

*Table 2. Examples of non-standard regional varieties in the Netlog Corpus.*

| Dialect | Netlog example | Standard Dutch | English |
|---|---|---|---|
| West-Flemish | zitn kik omeki zo verre? | Zit ik ineens zo ver? | Am I that far suddenly? |
| East-Flemish | est gedon | Is het gedaan? | Is it over/done? |
| Brabant | wa hedde gij die schoene gekocht | Waar heb jij die schoenen gekocht? | Where did you buy those shoes? |
| Limburg | hou gans veel van u | Ik hou heel veel van jou. | I love you very much. |

*Table 3. Examples of non-standard chat varieties in the Netlog Corpus.*

| Variation type | Netlog example | Standard Dutch | English |
|---|---|---|---|
| Omission of letters or words | kbda nimr | Ik heb dat niet meer. | I don't have that anymore. |
| Abbreviations | wrm<br>W8 | waarom<br>wacht | why<br>wait |
| Acronyms | hjg | hou je goed | take care |
| Character flooding | keiii mooiii | heel mooi | very beautiful |
| Concatenation | IkKanOokNiiiZonderU! | Ik kan ook niet zonder jou! | I can't live without you either! |

The following section provides an overview of the related research and discusses the compilation of a balanced subset of the Netlog corpus.

## Context

The study of the variation of linguistic characteristics in CMC according to an authors' age or gender has been a popular subject in both sociolinguistic and computational linguistic studies. At first, most of the studies were based on large collections of weblogs or "blogs", i.e., personal, informal writings listed in reverse-chronological order on a blogger's web site (e.g., Sarawgi, Gajulapalli & Choi, 2011; Mukherjee & Liu, 2010; Zang & Zang, 2010; Goswami, Sarkar & Rustagi, 2009; Argamon, Koppel, Pennebaker & Schler, 2007; Nowson & Oberlander, 2007; Yan & Yan, 2006). The main advantage of using blog corpora is that blog sites are publicly available and they usually contain information about the blogger's profile. Argamon et al. (2007), Goswami et al. (2009) already analysed distributions of (non-dictionary) words in blogs with regard to age and gender. Although their research focused on automatically predicting age and gender in CMC and not on linguistic variation, the study did report that teenage bloggers tend to use more non-dictionary words than adult bloggers do. Additionally, in their analysis of the number of non-dictionary words used per 1000 words across the 10s, 20s, 30s and higher age groups, female teen bloggers clearly used more non-dictionary words than the male teens, but the other age groups did not show significant gender differences. However, because other modes like online social networks or chat rooms usually do not provide open access to their users' profile, to our knowledge, there is only one study that employs a quantitative approach to investigate the correlation between the non-standard language usage and profile metadata in these modes of CMC: in his study on code choice and code-switching in Swiss-German chat rooms, Siebenhaar (2006) found that younger and older chatters use more dialect than chatters of the middle-aged group, but his study was only based on a list of 70 Standard German words that had a corresponding form in Swiss German. Consequently he could only investigate about 10 percent of the words in his chat corpus.

With regard to Flemish spoken discourse, Plevoets (2008) also studied the effects of age, gender and region in a corpus of spoken Dutch transcriptions (the CGN[3], which also includes spontaneous informal speech) on the use of "in-between"-varieties. His study revealed a correlation between register and region and between register and age, meaning that the dialectic background of the regions is still reflected in informal speech and that Flemish people born in 1940 or later clearly use non-standard variation in these situations. No correlation was found between register and gender. However, his study only included a prototypical set of linguistics features, the selection of which was determined by prior knowledge about the central Brabantic regiolect. Moreover, during the compilation of this corpus the informants who participated in the so-called 'spontaneous conversations' all received the explicit request to stick to Standard Dutch, which — as we mentioned before — does not correspond with actual Flemish colloquial speech practices. Furthermore, although there were no researchers present during the conversations, the informants actually had to record them. Both these factors, of course, did not contribute to the spontaneity of the conversations.

The present study examines the effects of age, gender and region — including potential interactions — on the use of typical chatspeak features and regional speech features in a corpus of Flemish social network posts. Moreover, we include all occurrences of non-standard language use in our analyses, which enables a more systematic study of both feature types. In the first part of our study, we go into the extraction and pre-processing of a subset of the Netlog corpus and discuss the balancing of our data over age, gender and region. We also describe the operationalization of non-standard language and the parameters that were used to categorise each non-standard word into one of both types. In the second part, we turn to a discussion of the results of the statistical analyses.

## Compilation of a Netlog Subset

Posts on the Netlog social network can contain multiple quotes from previous posts, of which we do not have the correct age, region and gender metadata. Therefore, the first step in pre-processing the data consisted of extracting only the last post of each interaction, of which we do have the required metadata, and saving these as separate files. In the second step of pre-processing we tokenized our dataset, lowercased all words and removed all punctuation marks, emoticons, e-mail addresses, phone numbers and hyperlinks so that only the word forms remained. Additionally, we reduced all four or more consecutive identical characters (character flooding) to three, so that e.g., the tokens 'niiice' and 'niiiiiice' (*nice*) were considered as the same type[4]. Given that the individual language usage of a Netlog user could bias our results if he or she was represented in the dataset by multiple posts, we chose to include only one post per user. Furthermore, we only selected posts that contained a minimum of three words.

For the age distribution in our dataset, we aimed at an equal distribution over 10s, 20s, 30s and 40s, but as the minimum age to create a Netlog profile is 13, we balanced our dataset per age year from 13 to 49. We did not include data from users older than 50, because there were not enough data available for each combination of age group and region in our dataset.

The smallest group of available authors in the corpus was the 34-year-old female group from Limburg with only 27 unique authors that had produced a post of at least three words. Therefore, we balanced our dataset as follows: 27 posts per age year (13 to 49), gender (male and female) and dialect region (West-Flanders, East-Flanders, Brabant and Limburg). This resulted in 7,952 posts (3,963 female and 3,989 male), which together accounted for 129,358 words. There were 1,998 authors from West-Flanders, 1,998 from East-Flanders, 1,995 authors from Brabant and 1,961 from Limburg.

The next phase of pre-processing consisted of extracting the number of standard and non-standard words per post. This categorization was done automatically by matching each word to the official wordlist of the Dutch Language[5], Standard English or Standard French[6]. We define a non-standard word as a word that does not occur in these lists, nor in a list of personal, geographical and brand names we collected manually. This resulted in the following information for each word in the subset: author ID, age, gender and region and a binary score for 'non-standardness': 0 if it was found in our standard word list or in the list of names and foreign words, and 1, otherwise. The dataset contained 115,914 (90%) standard and 13,444 non-standard words.

## Categorizing the Data

The Netlog corpus and its subset contain two types of non-standard language use: non-standard chatspeak forms and written representations of regional speech (either dialect or levelled dialect/regiolect) that are typical for everyday speech in Flanders. To investigate whether age, gender and/or region show a significant effect on the production of one of these types or both, we categorized each word that was labelled as non-standard during the pre-processing of the subset (see the previous section) as either chatspeak or regional.

First, we categorized all occurrences of spelling deviations and typing errors (i.e., both deliberate deviations from the standard language and undeliberate errors, e.g., 'vandaga' instead of 'vandaag' (*today*), 'gecontacteert' instead of 'gecontacteerd' (*contacted*)), character flooding (e.g., 'waaarom' instead of 'waarom' (*why*)) and all abbreviations and other creative adaptations that do not show any regional influence (e.g., 'bff' (*best friend forever*) and 'vr' for 'voor' (*for*)) as chat language. In the chat category we also included all forms that represented standard spoken Dutch, but were written in a non-standard way (e.g., all clitics without regional features like 'kheb' for "'k heb" (*I have*), standard Dutch pronunciation assimilations like 'feesje' instead of the standard written form 'feestje' (*party*) and reduced forms of personal pronouns such as 'dachtek' for "dacht ik*"* (*thought I*)). Non-standard English forms (e.g., 'disign' instead of 'design', 'wanna' for 'want to') were also included in the chat category.

Words in the regional speech category had to display dialect or regiolect features at the level of vocabulary, the representation of vowels and consonants, inflections or conjugations. In other words, although our analysis was performed at the lexeme level, phonological, morphological and lexical deviations from Standard Dutch were taken into account when categorizing a word into the regional speech category. In addition, abbreviated forms that showed regional influence in the process of abbreviating were also included. For example, the Standard Dutch word 'voor' (*for*) was represented by 'vr', 'veu' and 'vo(e)' in our subset: the first abbreviation 'vr' does not show any regional influence and was categorized as chat, while the other two varieties lean towards the Brabantic and West-Flemish pronunciation respectively and were therefore categorized as regional speech. We provide some examples of regional words, their Standard Dutch equivalents and their English translation in Table 4.

In the next two sections we examine the effects of age, gender and region on the distribution of both types of non-standard words (chatspeak and regional speech) in our Netlog subset. To assess these effects, we employed two separate mixed-effects logistic regression analyses with random by-author intercepts. In the first part we compare typical chatspeak productions with standard language productions and in the second part we compare regional word productions with standard language productions. Our analyses were based on 114,327 standard, 5,885 regional and 9,146 chatspeak features.

*Table 4. Examples of Flemish dialect variation in the Netlog subset.*

| Type of Dialect Variation | Netlog example | Standard Dutch | English |
|---|---|---|---|
| Vocabulary | *begaaie* | *vuil maken* | *to smudge* |
| | *kozn* | *neef* | *nephew* |
| | *lekstok* | *lolly* | *lollipop* |
| | *mokkes* | *meisjes* | *girls* |
| Vocals | *oltid* | *altijd* | *always* |
| | *woroem* | *waarom* | *why* |
| | *zuizu* | *sowieso* | *anyway* |
| | *weurst* | *worst* | *sausage* |
| Consonants | *skone* | *schone* | *beautiful* |
| | *vinne* | *vinden* | *to find* |
| | *percies* | *precies* | *just* |
| | *geleje* | *geleden* | *ago* |
| Inflections | *schatteke* | *schatje* | *honey* |
| | *dieje* | *die* | *that* |
| | *broere* | *broer* | *brother* |
| | *dadde* | *dat* | *that* |
| Conjugations | *kgaan* | *'k ga* | *I go* |
| | *hemk* | *heb 'k* | *have I* |
| | *eje* | *heb je* | *have you* |
| | *kzin* | *'k ben* | *I am* |

**Regiolectal Features Versus Chatspeak Features**

**Analysis 1: Effects of Age, Gender and Region on the Production of Chatspeak Features.**

To assess the effects of age, region, and gender and potential interactions on the probability of chatspeak production (chatspeak vs. standard), we employed stepwise mixed-effects logistic regression (Breslow & Clayton, 1993).[7] To allow for non-linearity in the effect of age, restricted cubic splines were used (Harrell, 2001), meaning that the age variable was split in a number of intervals with the effect of age on the probability of producing a chat word being fitted as a cubic curve within each interval. At the meeting points of these intervals — the knots — the curves are constrained to have smooth transitions, and beyond the observed values, the independent variable (in this case age) is assumed to be linearly related to the dependent variable (in this case chatspeak probability). Each knot adds a coefficient to the regression model, which can be tested for its statistical significance. There were 123,473 words in this data set: 114,327 (93%) standard words and 9,146 chat words. The observations were clustered in 7,950 authors (3,962 female, 3,988 male). The dataset included 1,994 authors from Brabant, 1,961 from Limburg, 1,997 from East-Flanders and 1,998 from West-Flanders.

To investigate the age effect, knots were placed at the begin- and endpoint, respectively 13 and 49, and at the ages 27, 33, and 39. The knots at 27, 33 and 39 were added after visual inspection of the mean proportions of non-standard language per age. Additionally, we also placed knots at the ages of 15 and 17 to test whether the Adolescent Peak Principle (e.g., Chambers, 2003; Labov, 2001) can also be found in our CMC data. Given that recent studies in Flemish linguistics (e.g., Vandekerckhove, 2009; Taeldeman, 2008; Plevoets, 2008; Van Keymeulen, 1993) discuss (and in some cases question) the importance of the Brabant region in the recent development of regiolects that have a wider geographical range than the local dialects do, in this analysis, region was treatment-coded with Brabant as the reference level.

The results of our analysis are summarized in Figure 2, which shows the effects of age, region and gender on the chat word probability. The final model contained significant main fixed effects for age ($\chi^2(6) = 1038.7$, $p <$

.001) and region ($\chi^2(3) = 25.1$, $p < .001$) with random intercepts for author (SD = 0.86). Gender did not have a significant effect ($\chi^2(1) = 2.58$, $p = .108$) and there were no significant interactions between age and region ($\chi^2(18) = 8.25$, $p = .975$), age and gender ($\chi^2(6) = 8.79$, $p = .186$) or gender and region ($\chi^2(3) = 3.75$, $p = .289$). As our analysis shows, age did have a significant non-linear effect on the chat word probability. The probability of the production of a chatspeak feature by users from Brabant rose from around 0.15 at 13 to a peak of 0.19 at 15, decreased sharply after that peak until it reached 0.05 at the age of 28. This rise and fall of the chat word probability strongly supports the Adolescent Peak Principle. Although female adolescents are thought to be the innovators of newly incoming non-standard forms — which in this case would be typical chat words like 'zjg' for 'zie je graag' (*love you*) — there was no interaction between age and gender, which means that our results show no change in the gender effect on the chat word probability during the adolescent peak. Quite unexpectedly, we found a very slight (and not significant) increase in the chat word probability between 29 and 33, but after the age of 33, it steadily decreases to reach 0.03 at the age of 41 and remains constant at older ages. As for the region variable, we found no significant difference in chat word probability between East-Flanders and Brabant ($\beta = 0.05$, $SE = 0.05$, $p = .288$), but Netlog users from Limburg were significantly less likely to produce chatspeak features than users from Brabant ($\beta = -0.11$, $SE = 0.05$, $p = .023$), while users from West-Flanders employed significantly more chatspeak forms than users from Brabant ($\beta = 0.12$, $SE = 0.05$, $p = .010$).

*Figure 2. Estimated effects of age and region on the probability that chat speak forms are produced. The curves represent restricted cubic splines with knots placed at the arrows on the* x-*axis.*

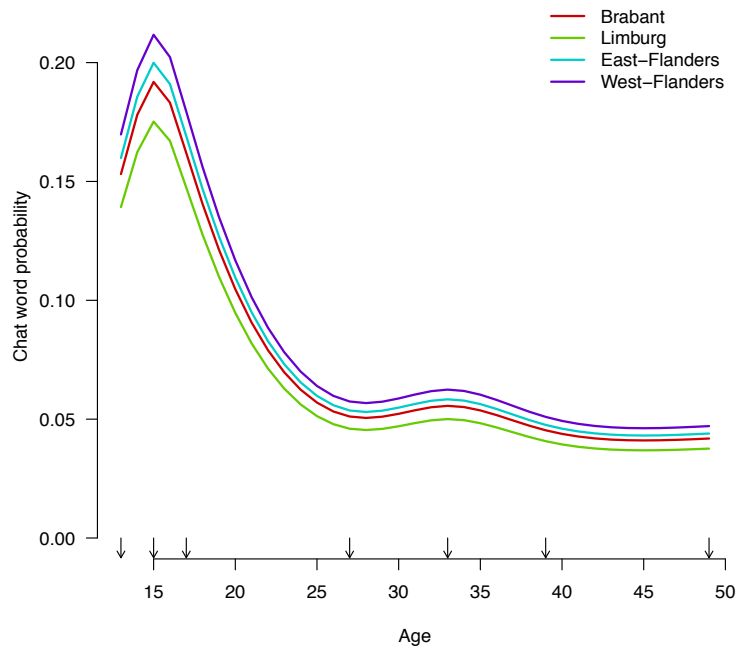

**Analysis 2: Effects of Age, Gender and Region on the Production of Regiolectal Features**

To assess the effects of age, region, and gender and potential interactions on the probability of a regional word production (regional speech vs. standard), we used the same approach as for the previous analysis, namely forward stepwise mixed-effects logistic regression. The data set consisted of 120,212 words, of which 114,327 were standard words and 5,885 were regional speech features.[8] These observations were clustered in 7,932 authors (3,953 female, 3,979 male). In this analysis, the dataset contained Netlog posts from 1,992 Brabantic, 1,956 Limburgish, 1,988 East-Flemish and 1,996 West-Flemish authors. To capture the non-linearity in the relation between age and dialect word probability, a restricted cubic spline function was fitted with five knots. After visual inspection of the regional word production proportions per author, knots were placed at the endpoints 13 and 49, and at the ages 15, 17, and 33. The knots at 15 and 17 again tested the adolescent peak hypothesis. There were no additional non-linearities to account for ($\chi^2(1) = 0.18$, $p = .667$).

The results of the regional speech use analysis are summarized in Figure 3. The analysis showed significant main fixed effects for age ($\chi^2(4) = 782.36$, $p < .001$) and region ($\chi^2(3) = 98.12$, $p < .001$) with random intercepts for author (SD = 1.30). Gender again did not show a significant effect on the regional word probability ($\chi^2(1) = 0.64$, $p = .425$), and there were no significant interactions between gender and age ($\chi^2(4) = 8.89$, $p = .064$), or gender and region ($\chi^2(3) = 1.55$, $p = .670$). Although a likelihood ratio test indicated that there was a significant interaction between age and region ($\chi^2(12) = 33.04$, $p < .001$), none of the interaction coefficients differed significantly from zero (all Wald $z$ $p$-values > .16). With regard to the age variable, as is visualized in Figure 3, the probability of producing a regional word for authors from Brabant rose from around 0.07 at the age of 13 to a peak of 0.11 at 15, and then decreased smoothly to a value around 0.01 at 49. As for the region variable, we found no significant difference in regional word probability between East-Flanders and Brabant ($\beta = -0.01$, $SE = 0.07$, $p = .875$). The curves for these two regions coincide almost perfectly. However, Netlog users from Limburg were significantly less likely to produce regional speech features than users from Brabant ($\beta = -0.42$, $SE = 0.07$, $p < .001$). Additionally, West-Flemish authors employed significantly more regional words than users from Brabant did ($\beta = 0.26$, $SE = 0.07$, $p < .001$).

*Figure 3. Estimated effects of age and region on the probability that regional and/or dialect words are produced. The curves represent restricted cubic splines with knots placed at the arrows on the* x-*axis.*

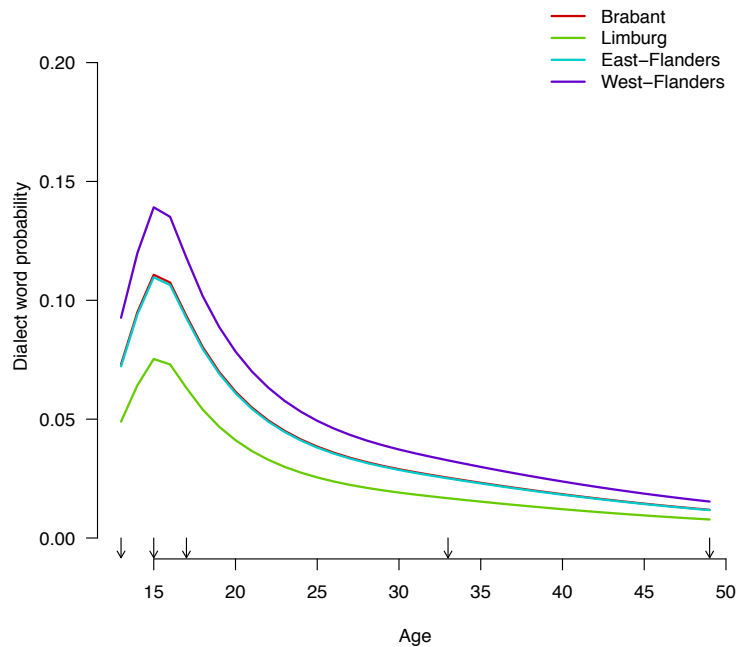

Our results show that, in spite of the ongoing process of dialect loss in Flanders, especially among young people (e.g., Vandekerckhove, 2009), the curve's rise and fall around the age of 15 also confirms the presence of the Adolescent Peak Principle in the use of regional speech forms in Flemish CMC. These findings could be explained by Siebenhaar's (2006) theory that, although their knowledge of their local dialect is declining, using dialect forms in CMC is attractive for teenagers, because the writing process is dissociated from orthography and other writing standards that are learnt in a school context. Moreover, the effects we found regarding the region variable can be aligned with the "extraordinary dialect vitality" that has been ascribed to the West-Flemish region (compared to other regions in Dutch-speaking Belgium), while in Limburg, the process of dialect loss was accelerated by the industrialization of some parts of the province which led to massive immigration from other parts of Belgium and from foreign countries (e.g., Vandekerckhove, 2009; Belemans & Keulen, 2004; Belemans, 1997; Van Keymeulen, 1993).

**Discussion**

In this study, we used a subset of the Netlog corpus that was balanced according to age (13 to 49), gender and dialect region (West-Flanders, East-Flanders, Brabant and Limburg) to assess the effects of these sociological variables on non-standard linguistic variation in Flemish social network posts. What makes this corpus interesting for linguistic variation studies is that it contains two types of non-standard language use: newly incoming non-standard forms that are inherent to the medium and written representations of older regional and dialect forms that are typical for colloquial speech. Therefore, we manually categorized each non-standard word into either 'chatspeak' or 'regional speech' and applied a forward stepwise mixed-effects logistic regression analysis to examine the effects of age, gender and region on the production of both chatspeak and regiolectal forms in our subset. Our analyses showed that age had a significant non-linear effect on the production of both types, indicating that the principle of Age Grading, that was reported in previous spoken discourse studies of e.g., Downes (1984), Holmes (1992) and Labov (1994), is reflected in Flemish CMC, if only partially, because we were not able to include data from the older (plus 50) age groups in the subset.

The pattern of Age Grading is usually found to be a U-shaped curve (Downes, 1984) in which the use of non-standard linguistic variables tends to peak during adolescence. Our analyses indeed show that both the chat word probability and the regional word probability peak between the ages of 13 and 15, which corroborates the presence of the Adolescent Peak Principle. The latter finding is quite remarkable given the on-going dialect loss processes in Flanders, which mainly affect these younger groups. However, except for West-Flanders, most adolescents produce regional speech with a wide geographical reach rather than small-scale local dialect forms. The latter may not be part of their verbal repertoire anymore.  In spite of that, they clearly seem to enjoy displaying their (limited) knowledge of regional varieties, the use of which appears to function as a (regional) solidarity marker.  Moreover, producing non-standard speech in a written medium dissociates the writing process from orthography and writing standards learnt at school, which makes it extra attractive (Siebenhaar, 2006). Yet, this still does not explain why the peak of the regional word probability starts to decrease after the age of fifteen. Flemish adolescents surely do not possess a more extensive knowledge of regional speech than post-adolescents do and post-adolescents are equally familiar with writing non-standard forms as their younger peers. The presence of an Adolescent Peak in the regional word probability could therefore indicate that the use of regional words in chat becomes less 'cool' during post-adolescence, because it is associated with a younger age group. However, this hypothesis requires further research to be confirmed.

In accordance with Plevoets' (2008) study, we found no significant effect for gender. Additionally, although female adolescents are thought to be the innovators of newly incoming non-standard (in this case chatspeak) forms, our results suggested no change in the gender effect on the probability that chatspeak features are produced during the adolescent peak.

With regard to the region effect, our findings support previous dialectological studies stating that the West-Flemish region appears to be marked by extraordinary dialect preservation, while Limburg is subject to dramatic dialect loss, but our study also adds a new element: our analyses show the same interregional differences for the use of typical chatspeak features, which are non-standard but not rooted in the local dialects. These results suggest that the degree of dialect vitality in a region can function as a threshold, not only for including regiolectal features, but also for including other non-standard (chat) features in chatspeak.

Summarizing, our study does not only confirm that combining quantitative and qualitative approaches is essential when dealing with non-standard linguistic variation in a CMC corpus (e.g., Vandekerckhove & Nobels, 2010; Siebenhaar, 2006; Androutsopoulos & Ziegler, 2004), it also presents a corpus-based approach which allows to shift the research focus from a selection of non-standard linguistic variables, leading to a limited view of non-standard language variation, to a more systematic approach which incorporates all non-standard lexemes in the selected subcorpus.  Until now, this kind of inclusive approach seems to have been absent in the research on online social network discourse (Herring et al., 2013). The most striking findings of the present study relate, first of all, to the unambiguous illustration of the Adolescent Peak Principle and, secondly, to the intriguing co-occurrence of non-standard regional features and non-standard chatspeak features in written social network posts produced by Flemish teenagers. The latter finding offers new perspectives on the notion of non-standard speech and demonstrates the added value of CMC research for variational linguistics.

---

NOTES

[1] http://nl.netlog.com/

[2] In Standard Dutch, 'schoon' means *clean*.

[3] http://lands.let.kun.nl/cgn/

[4] We did not reduce them to two, because then the word could become standard Dutch. For example, the intensifier in 'zooo slecht' (*sooo bad*) could be wrongly interpreted as the standard form 'zoo' (*idem*).

[5] The *Bronbestand van de woordenlijst van de Nederlandse taal* was used to create the official Dutch spelling dictionary and is available via the TST-Centrale website: http://www.inl.nl/tst-centrale/nl/producten/lexica/bronbestand-woordenlijst-nederlandse-taal-2005/7-26.

[6] http://packages.debian.org/nl/lenny/

[7] We used the R package *lme4* (Bates, Mächler, & Bolker, 2012) for all mixed-effects regression analyses in this paper.

[8] The standard word observations are the same as the standard words in the chatspeak analysis.